\documentclass{article}

\newif\ifanon
\anonfalse
\ifanon\usepackage[dblblindworkshop]{neurips_2026}\else\usepackage[preprint]{neurips_2026}\fi

\workshoptitle{WORKSHOP TITLE}

\usepackage[utf8]{inputenc} 
\usepackage[T1]{fontenc}    
\usepackage{hyperref}       
\usepackage{url}            
\usepackage{booktabs}       
\usepackage{amsfonts}       
\usepackage{nicefrac}       
\usepackage{microtype}      
\usepackage[table]{xcolor}  
\usepackage{makecell}       
\usepackage{graphicx}       
\usepackage{flafter}        
\usepackage{needspace}      

\title{Agents unlock new capabilities through Switching LoRA Adapters as a Tool (SLAaaT)}

\author{%
  Kenneth Ge \\
  Independent\\
  San Francisco, CA, United States of America\\
  \texttt{kennethkouge@gmail.com}
}

\begin{document}

\maketitle

\begin{abstract}
  Post-training can unlock new capabilities and improve performance on specialized tasks, but sometimes at the cost of catastrophic forgetting in other domains. This poses a problem in long agent trajectories that compose different capabilities. We reject this tradeoff by giving an agent a tool to switch between specialized LoRA adapters mid-trace. To test its effectiveness, we compose two synthetic coding tasks that are logically simple but require specialization. We find that this allows the model to solve problems it previously could not, that the model is able to switch autonomously (and find a new strategy that beats our human heuristic baseline on one task), and that this incurs an up to an 18x reduction in capability tax compared to an agent using only one specialized adapter. Our approach also substantially outperforms spawning subagents in both task capabilities (solving 4 of our hardest tasks versus none) and token usage (46.1x fewer tokens in some scenarios).
\end{abstract}

\section{Introduction}
\label{sec:intro}

\begin{figure}[t]
  \centering
  \includegraphics[width=\linewidth]{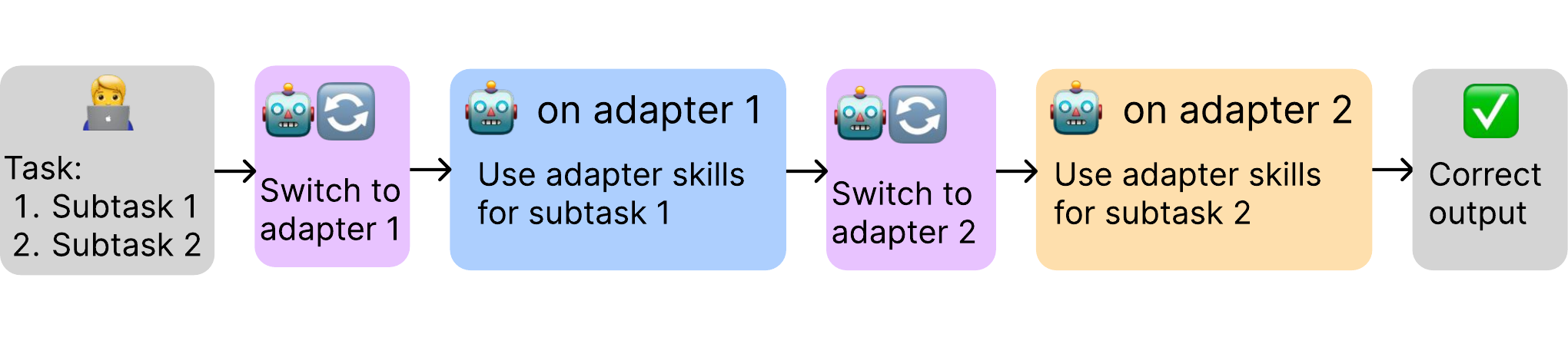}
  \caption{The agent switches its own LoRA adapter mid-trajectory, activating the right specialist for each subtask.}
  \label{fig:overview}
\end{figure}

\begin{table}[t]
  \centering
  \caption{Task passes (out of 50 per cell) across the four notation bindings. Shading scales with pass rate. Rows group into single-configuration arms, routed arms over the fixed \{fauxjson, fauxthon\} library, and the frontier reference.}
  \label{tab:main-grid}
  \small
  \setlength{\tabcolsep}{5pt}
  \begin{tabular}{l c c c c c}
    \toprule
    Condition                    & \shortstack{json--python\\(control)}          & fauxjson--python                              & json--fauxthon                                & \shortstack{keystone\\(both)} & Total /200                  \\
    \midrule
    base                         & \cellcolor[HTML]{5EBA9C}\textcolor{white}{42} & \cellcolor[HTML]{FFFFFF}0                     & \cellcolor[HTML]{AADBCB}22                    & \cellcolor[HTML]{FFFFFF}0     & \cellcolor[HTML]{C2E5D9}64  \\
    fauxjson (pinned)            & \cellcolor[HTML]{FFFFFF}0                     & \cellcolor[HTML]{FFFFFF}0                     & \cellcolor[HTML]{FFFFFF}0                     & \cellcolor[HTML]{FFFFFF}0     & \cellcolor[HTML]{FFFFFF}0   \\
    fauxthon (pinned)            & \cellcolor[HTML]{B6E0D2}19                    & \cellcolor[HTML]{FFFFFF}0                     & \cellcolor[HTML]{D5EDE5}11                    & \cellcolor[HTML]{FFFFFF}0     & \cellcolor[HTML]{E2F3ED}30  \\
    fused                        & \cellcolor[HTML]{FFFFFF}0                     & \cellcolor[HTML]{FFFFFF}0                     & \cellcolor[HTML]{FFFFFF}0                     & \cellcolor[HTML]{FFFFFF}0     & \cellcolor[HTML]{FFFFFF}0   \\
    \midrule arrow (token-level) & \cellcolor[HTML]{61BB9F}\textcolor{white}{41} & \cellcolor[HTML]{FFFFFF}0                     & \cellcolor[HTML]{F7FCFA}2                     & \cellcolor[HTML]{FFFFFF}0     & \cellcolor[HTML]{D6EDE6}43  \\
    subagent                     & \cellcolor[HTML]{6DC0A6}\textcolor{white}{38} & \cellcolor[HTML]{F7FCFA}2                     & \cellcolor[HTML]{B6E0D2}19                    & \cellcolor[HTML]{FFFFFF}0     & \cellcolor[HTML]{C6E7DC}59  \\
    heuristic switch             & \cellcolor[HTML]{52B595}\textcolor{white}{45} & \cellcolor[HTML]{BAE1D5}18                    & \cellcolor[HTML]{8CCEB9}\textcolor{white}{30} & \cellcolor[HTML]{E0F2EC}8     & \cellcolor[HTML]{9ED5C4}101 \\
    autoswitch (model)           & \cellcolor[HTML]{52B595}\textcolor{white}{45} & \cellcolor[HTML]{71C2A8}\textcolor{white}{37} & \cellcolor[HTML]{8CCEB9}\textcolor{white}{30} & \cellcolor[HTML]{F0F8F6}4     & \cellcolor[HTML]{90CFBB}116 \\
    \midrule sonnet5 (reference) & \cellcolor[HTML]{43AE8C}\textcolor{white}{49} & \cellcolor[HTML]{E0F2EC}8                     & \cellcolor[HTML]{C2E5D9}16                    & \cellcolor[HTML]{F7FCFA}2     & \cellcolor[HTML]{B7E0D3}75  \\
    \bottomrule
  \end{tabular}
\end{table}

For a given architecture and parameter budget, models often have to make tradeoffs between specialization and generality \citep{yue2025does}. This is especially pronounced in small language models, which readily absorb new capabilities \citep{fu2023specializing} but can forget old ones \citep{marek2026forgetting}.

Fine-tuning lets small models match massive ones on specific tasks \citep{patil2023gorilla}. However, such models are currently used as subagent specialists with limited functionality \citep{belcak2025small}. Even when exposed as tools in an agentic setting, these specialist models are invoked and switched to once \citep{shekar2025adaptive}.

In this paper, we allow an agent to repeatedly hotswap its own LoRA adapters within the same trajectory (Figure~\ref{fig:overview}). We find that the agent composes these swaps to solve new synthetic tasks without sacrificing overall capacity. With just one in-context example totaling 103 tokens, the model was able to learn when to use this tool, nearing the performance of our human heuristic switching on most tasks and exceeding on one. The model outperformed all other baselines, including staying on a single LoRA adapter, using a fused adapter fine-tuned on all tasks, Arrow \citep{ostapenko2024towards}, and invoking subagents as a tool. Table~\ref{tab:main-grid} summarizes our results. \ifanon\else All code, prompts, and data are available on Huggingface at \url{https://huggingface.co/kennethge123/autolora}.\fi

\needspace{16\baselineskip}
\section{Method}
\label{sec:method}

\subsection{Task Design}
\label{sec:task-design}
A task that isolates whether post-trained skills compose should have the following properties:
\begin{itemize}
  \item The necessary skills are absent from the base model and nontrivial to elicit via in-context examples (the capability is gained via post-training)
  \item The task decomposes into multiple stages that each require one skill, and success in a later stage is conditional upon success in an earlier one
  \item The subtasks are intrinsically easy, with equivalent counterparts that are already solvable by the model, so differences in performance are not about difficulty
  \item Task success is verifiable
  \item Training corpora for each subtask are fully disjoint
\end{itemize}

We designed a synthetic data processing pipeline with two steps: translating data from YAML to either JSON or a synthetic markup format called Fauxjson, and writing a program in either Python or Fauxthon to process this data and output the result as XML. Fauxjson and Fauxthon are just JSON and Python with their keywords/symbols cycled. For example, the symbols [\texttt{\{}, \texttt{\}}, \texttt{:}] might map to [\texttt{\}}, \texttt{:}, \texttt{\{}]. This shares the tokenizer, interferes with the model's strong priors, and keeps the spec simple.

In total, our test suite included 5 different processing task types (lossless conversion, counting, aggregation, sorting, and grouping), with 10 different tasks per task type and 10 YAML inputs each (1 given to the model, 9 held out). We scored a 1 if the translated data file was correct and if the program produced a correct result on all 10 inputs, and 0 otherwise.

\subsection{Adapter Design}
\label{sec:adapter-design}
Our fine-tuning data consisted of 20k synthetic YAML to Fauxjson translation pairs and 20k Fauxthon coding exercises transpiled from MBPP \citep{austin2021program} (950 rows), Magicoder \citep{wei2024magicoder} (9,050 rows), and synthetic programs (10k rows). All synthetic data was either transpiled programmatically or generated by Claude Sonnet 5; Fauxjson and Fauxthon were generated by our transpiler and verified to round-trip. Training details are in Appendix~\ref{app:training}.

The Fauxjson, Fauxthon, and fused adapters were each trained on 1 epoch. The fused adapter used a shuffled combination of both datasets.

\subsection{Model and Sampling}
\label{sec:model-sampling}
We used Qwen3.6-35B-A3B, a model large enough to understand tool calls and solve non-trivial coding problems but still capacity constrained. For sampling, we gave the model 20 turns and a 4096 token budget, and we human-validated that these limits were only reached when models were stuck generating repetitive content. We kept thinking off.

\subsection{Adapter Invocation and Tools}
\label{sec:invocation-tools}
We used a standard subagent invocation tool where the host model writes a prompt. For our heuristic switching decisions, we monitored for \texttt{file\_write} tools and switched to the adapter corresponding to the file extension. Autoswitch is a new tool where the model selects either base (no adapter), fauxthon, or fauxjson, and the harness switches the model accordingly.

\section{Results}
\label{sec:experiments}

We answer three questions: do separately trained skills compose, can the model route them itself, and what does each configuration cost?

Table~\ref{tab:main-grid} summarizes end-to-end pass rates for every condition, and Figure~\ref{fig:retention} shows how much base capability each retains. Statistical notes in Appendix~\ref{app:stats}.

\begin{figure}[t]
  \centering
  \includegraphics[width=0.85\linewidth]{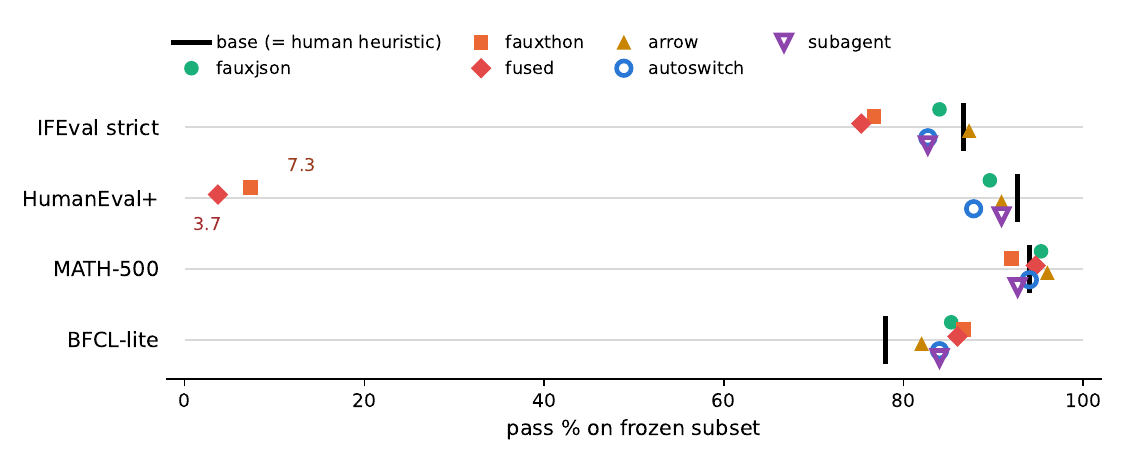}
  \caption{Retention of general capabilities. Pass rate on random frozen subsets of IFEval, HumanEval+, MATH-500, and BFCL-lite for each condition; the black bar marks the base model (equivalently, the human heuristic). The fused and fauxthon adapters pay a large capability tax, most severely on HumanEval+ (3.7\% and 7.3\%), whereas the routed conditions (autoswitch, subagent, arrow) stay close to base. This is the ``capability tax'' referred to throughout Section~\ref{sec:experiments}.}
  \label{fig:retention}
\end{figure}

\definecolor{passgreen}{RGB}{27,120,88}

\begin{table}[h]
  \centering
  \caption{Testing our four single-configuration conditions on just fauxjson and fauxthon tasks outside of agentic tool calling confounds. Shading scales with score (higher is better).}
  \label{tab:trial5-probe}
  \begin{tabular}{lcccc}
    \toprule
    Subtask                     & base                     & fauxjson                   & fauxthon                                       & fused                                          \\
    \midrule
    fauxjson translation /50    & \cellcolor{passgreen!0}0 & \cellcolor{passgreen!40}23 & \cellcolor{passgreen!0}0                       & \cellcolor{passgreen!70}\textcolor{white}{44}  \\
    fauxthon program tests /500 & \cellcolor{passgreen!0}0 & \cellcolor{passgreen!0}0   & \cellcolor{passgreen!55}\textcolor{white}{290} & \cellcolor{passgreen!70}\textcolor{white}{360} \\
    \bottomrule
  \end{tabular}
\end{table}

\paragraph{Composition} We first verified that skills live in their respective adapters. We reran each of our four single-adapter configurations on just the Fauxjson translation tasks and the Fauxthon programming tasks. To remove the agentic harness confound, we had the model just emit a codeblock. We find that each adapter performs well at its respective task, and that the fused adapter actually performs by far the best on knowledge for both tasks (Table~\ref{tab:trial5-probe}). However, fusion fails all agentic tasks (Table~\ref{tab:main-grid}). Analyzing the traces, we find strong evidence that this is due to catastrophic forgetting: the fusion condition forgets Python and JSON, and meanwhile writes syntactically correct but semantically incorrect Fauxthon code.

Ideally, composition allows the probability of end-to-end success to be directly related to success on each subtask. For the human heuristic condition, we find that this largely holds true: the keystone score of $8/50$ nears the predicted $\nicefrac{18}{50} \times \nicefrac{30}{50} \approx 11/50$, so the switching mechanism itself largely preserves continuity across the trace.

\paragraph{Routing} Autoswitch beats our heuristic on Fauxjson-Python because it invented an innovative new policy: using the Fauxjson adapter to write Python code. This suggests models can already autonomously discover optimal routing choices. Meanwhile, Autoswitch's actual score on the Fauxjson--Fauxthon (keystone) condition, $4/50$, is far lower than predicted by its performance on the individual Fauxjson--Python and JSON--Fauxthon tasks, $\nicefrac{37}{50} \times \nicefrac{30}{50} \approx 22/50$, suggesting large capability gains remain on the table.

\definecolor{tokred}{RGB}{215,48,39}

\begin{table}[h]
  \centering
  \caption{Output tokens per task attempt. Shading scales with token count (more tokens is worse).}
  \label{tab:token-stats}
  \begin{tabular}{lcccc}
    \toprule
    Condition               & \makecell{passed:\\mean}     & passed: median              & failed: mean                                     & failed: median                                   \\
    \midrule
    autoswitch (116 passes) & \cellcolor{tokred!8}1{,}793  & \cellcolor{tokred!4}834     & \cellcolor{tokred!10}2{,}138                     & \cellcolor{tokred!4}797                          \\
    subagent (59 passes)    & \cellcolor{tokred!24}5{,}242 & \cellcolor{tokred!5}1{,}071 & \cellcolor{tokred!75}\textcolor{white}{39{,}407} & \cellcolor{tokred!70}\textcolor{white}{36{,}731} \\
    \bottomrule
  \end{tabular}
\end{table}

\paragraph{Costs} Both the fused and fauxthon adapters incur a significant capability tax that switching largely avoids (Figure~\ref{fig:retention}). Compared to the subagent condition, Autoswitch incurred approximately the same capability tax, both from routing unnecessarily. Autoswitch also used 1.3x fewer tokens for the median success compared to subagent, and 46.1x fewer tokens for the median failure (Table~\ref{tab:token-stats}).

\section{Limitations and Future Directions}
\label{sec:future-directions}

\paragraph{Limitations} Our tasks are synthetic by design: Fauxjson and Fauxthon isolate the composition question. All results use a single base model (Qwen3.6-35B-A3B), a single training seed, and one sample per task, so we report binomial confidence intervals (Appendix~\ref{app:stats}). Our routing library contains two adapters and scaling to many adapters may elicit different behavior. Finally, the human heuristic is a fixed file-extension rule rather than an optimal policy, so ``beating the heuristic'' should be read as beating a strong hand-written rule, not an upper bound.

\paragraph{Future directions} This paper establishes that models can effectively switch their own LoRA adapters under policies that sometimes beat human switching decisions. Future work could use reinforcement learning to train an even stronger routing policy. The functionality already exists to reuse the KV cache for different LoRA adapters, solving the biggest technical challenge \citep{li2025efficient}. Future work could scale this method to production.

\begin{ack}
  Thank you to Thinking Machines for supporting this research with a Tinker Research Grant.
\end{ack}

\bibliographystyle{plainnat}
\bibliography{references}


\appendix

\section{Training and Evaluation Details}
\label{app:training}

All adapters were trained via the Tinker API on Qwen3.6-35B-A3B for one epoch.
\begin{table}[h]
  \centering
  \caption{LoRA training hyperparameters.}
  \label{tab:hparams}
  \begin{tabular}{ll}
    \toprule
    Base model             & Qwen3.6-35B-A3B                                        \\
    LoRA rank              & 16                                                     \\
    Learning rate          & $1\times10^{-4}$, 3\% linear warmup, cosine decay to 0 \\
    Optimizer              & Adam                                                   \\
    Batch size             & 128 conversations                                      \\
    Epochs                 & 1                                                      \\
    Fauxjson training rows & 20{,}000                                               \\
    Fauxthon training rows & 20{,}000                                               \\
    \bottomrule
  \end{tabular}
\end{table}
Evaluation used 20 turns and a 4{,}096-token budget per attempt with thinking disabled (Section~\ref{sec:model-sampling}). Test tasks are held out from all training data; each of the 50 tasks per condition uses 1 YAML input shown to the model and 9 held-out inputs for scoring.

\section{Compute}
\label{app:compute}
All training and evaluation ran on the Tinker API (Thinking Machines), except the Arrow baseline, which ran locally on a single NVIDIA DGX Spark. Producing the main results table (Table~\ref{tab:main-grid}) took roughly two days of wall-clock time end to end.

\section{Statistical Notes}
\label{app:stats}
Every cell in Table~\ref{tab:main-grid} is a pass count out of $n=50$ independent tasks from a single run (one seed=5, one sample per task). Treating each cell as a binomial proportion, the 95\% Wilson interval half-width is at most $\approx 14$ percentage points (at $p=0.5$) and shrinks toward the extremes (e.g.\ $45/50$: $[79\%, 96\%]$; $0/50$: $[0\%, 7\%]$). The conclusions we draw---fusion and pinned adapters failing agentic tasks outright, and autoswitch's total of 116 versus subagent's 59---are separated by margins well outside these intervals. We did not run multiple seeds because of compute cost.

\section{Assets and Licenses}
\label{app:assets}
We use Qwen3.6-35B-A3B (Apache-2.0), MBPP \citep{austin2021program} (CC-BY-4.0), and Magicoder \citep{wei2024magicoder} (MIT) as sources for transpiled training data.
Our released assets---adapters, transpiler, task suite, prompts, and evaluation harness---\ifanon\else are available at \url{https://huggingface.co/kennethge123/autolora}\fi{} with a README describing setup and reproduction.

\section{Broader Impact}
\label{app:impact}
This work makes it cheaper to give a small model many narrow skills without paying a general capability tax, which lowers the barrier to deploying capable agents on modest hardware. That same property could make it easier to bolt undesirable specializations onto an open model, though it does not enable anything a full fine-tune could not already do; our released adapters target synthetic languages with no real-world use and pose no incremental risk.

\section{AI Usage}
\label{app:ai-usage}
LLMs were used to help draft the appendix, format the \LaTeX{} source, write captions, and make minor edits to the text. LLMs did not affect the methodology, results, or conclusions.


\end{document}